# Predicting Local Climate Zones using Urban Morphometrics and Satellite Imagery

Hugo MAJER[a,*], Martin FLEISCHMANN[b]

[a] Department of Applied Geoinformatics and Cartography, Faculty of Science, Charles University, Albertov 6, 128 00 Prague, Czechia

[b] Department of Social Geography and Regional Development, Faculty of Science, Charles University, Albertov 6, 128 00 Prague, Czechia

[*] Corresponding author:
*e-mail address*: majerh@natur.cuni.cz (H. Majer)

**Abstract**

The Local Climate Zone (LCZ) framework is commonly employed to represent urban form in morphological analyses despite its mapping predominantly relies on satellite imagery. Urban morphometrics, describing urban form via numerical measures of physical aspects and spatial relationships of its elements, offers another avenue. This study evaluates the ability of morphometric assessment to predict LCZs using a) a morphometric-based LCZ prediction, and b) a fusion-based LCZ prediction combining morphometrics with satellite imagery. We calculate 321 2D morphometric attributes from building footprints and street networks, covering their various properties at multiple spatial scales. Subsequently, we develop four classification schemes: morphometric-based prediction, baseline image-based prediction, and two techniques fusing morphometrics with imagery. We evaluate them across five sites. Results from the morphometric-based prediction indicate that the correspondence between 2D urban morphometrics and urban LCZ types is selective and inconsistent, rendering the efficacy of this method site-dependent. Nevertheless, it demonstrated that a much broader range of urban form properties is relevant for distinguishing LCZ types compared to standard parameters. Relative to the image-based baseline, the fusion yielded relatively distinct accuracy improvements for urban LCZ types at two sites; however, gains at the remaining sites were negligible or even slightly negative, suggesting that the benefits of fusion are modest and inconsistent. Collectively, these results indicate that the relationship between the LCZs and the measurable, visible aspects of urban form is tenuous, thus the LCZ framework should be used with caution in morphological studies.

# 1. Introduction

The physical form of a city fundamentally influences its environmental and socioeconomic performance. The configuration and properties of urban elements, including buildings, streets, or neighborhoods, can be linked to urban flooding susceptibility (Wang et al., 2023), energy consumption (Haffner et al., 2024), and public health (Nieuwenhuijsen, 2021). Urban morphology investigates these relationships often using typologies comprising a limited number of distinct urban types, a simplification that facilitates systematic and comparable morphological analyses (Fleischmann et al., 2025). To derive those, urban morphology increasingly relies on quantitative methods to address scalability and replicability challenges (Porta et al., 2022; Oliveira, 2024). Among these, urban morphometrics, a numerical assessment of individual elements of urban form and their spatial relationship, is gaining prominence as a distinct approach to urban form characterization (Oliveira and Porta, 2025).

Despite advances in morphometrics, a widely used typology in morphological analyses is the Local Climate Zone (LCZ) framework (Taubenböck et al., 2020; Naserikia et al., 2023; Debray et al., 2025) despite its origins in urban climate research (Stewart and Oke, 2012). The scheme comprises ten urban types (LCZs 1–10) distinguished by building density (compact/open/sparse), height (low/mid/high), size (large/other), function (general/industrial), and material (heavy/lightweight), alongside seven natural types (LCZs A–G) capturing vegetation and land cover characteristics (Figure 1) (Taubenböck et al., 2020; Han et al., 2024).

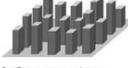

Figure 1. Local Climate Zones framework (Stewart and Oke, 2012).

Within the context of LCZ mapping, studies incorporating numerical urban form descriptors similar to urban morphometrics have emerged (Hidalgo et al., 2019; Yoo et al., 2020; Xu et al., 2023; Ao et al., 2025), mainly driven by the demand for novel approaches addressing persisting mapping efficiency and accuracy concerns (Han et al., 2024; Acosta-Fernandéz et al., 2025). However, these efforts relied on a small set of specific and isolated metrics providing, at best, only limited description of urban form.

Here we present the most exhaustive application of urban morphometrics approach in the context of LCZ mapping. Based on the premise that LCZ urban types reflect different urban form configurations, urban morphometrics should therefore capture relevant features for LCZs differentiation and offer predictive value. After a brief review of LCZ mapping methods and prior attempts utilizing urban form descriptors, we propose and evaluate: 1) morphometric-based LCZ prediction; 2) fusion-based LCZ prediction combining morphometrics and satellite imagery.

## 2. Background

Currently, three approaches to LCZ mapping are mainstream: geographic information system (GIS)-based methods, remote sensing (RS)-based methods, and combined methods. GIS-based approaches determine LCZ types using defined thresholds for physical surface parameters, such as sky view factor, aspect ratio, and building surface fraction, often referred to as urban canopy parameters (Quan and Bansal, 2021). While capable of high accuracy, these methods demand extensive input data (Acosta-Fernandéz et al., 2025). Conversely, RS-based approaches are preferred due to the high availability, global coverage, and sufficient spatiotemporal resolution of satellite imagery, despite requiring RS expertise and reference datasets (Han et al., 2024). Here the supervised classification of imagery is utilized, pioneered by the WUDAPT workflow employing pixel-based Random Forest (RF) classification (Bechtel et al., 2015). Subsequently, scene-based classification using Convolutional Neural Networks (CNN) has become the dominant method (Huang et al., 2023) due to substantial improvements in accuracy (Rosentreter et al., 2020). Meanwhile, approaches combining GIS- and RS-based methods are the least used (Huang et al., 2023), despite potential to outperform single-source methods (Fonte et al., 2019). Two primary integration strategies exist: directly integrating urban canopy parameters or other auxiliary layers into image classification (Kim et al., 2021; Vavassori et al., 2024); or separately classifying urban LCZ types using GIS-based methods and natural LCZ types via RS-based methods (Zhou et al., 2021).

Focusing on prior attempts utilizing urban form descriptors for LCZ prediction, Yoo et al. (2020) combined Sentinel-2 imagery with building area and building height and reported inconsistent impacts on accuracy with both improvements and deteriorations compared to RS-based approach. While Geletič and Lehnert (2016) and Yang et al. (2021) incorporated building density into GIS-based workflows, Ao et al. (2025) combined it with Landsat 8 imagery, observing accuracy improvements across all urban LCZ types. Similarly, Xu et al. (2023) applied a GIS-based workflow utilizing building coverage area ratio within parcels. Adopting a topological perspective, Li et al. (2025) used a graph network approach to derive distances between a specific number of nearest neighboring buildings, integrating these alongside building area in a building-centric LCZ mapping method. Notably, Hidalgo et al. (2019) utilized the most extensive set of metrics within a GIS-based framework, specifically building density, coverage area ratio, and the mean and median of minimal interbuilding distances.

# 3. Methods

The method is structured into four parts (Figure 2). First, we measure morphometric attributes based on GIS input data. Those, together with satellite imagery, are classified using a set of classifiers and classification schemes, to be evaluated and compared. The following sections outline each of these steps.

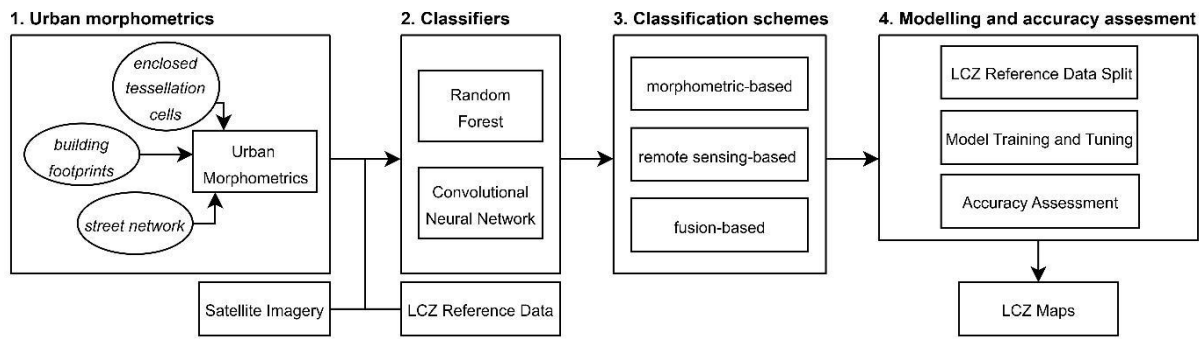

Figure 2. Illustration of the method.

## 3.1. Urban morphometrics

We calculate urban morphometric attributes using the *momepy* Python library (Fleischmann, 2019), based on building footprints, street network, and enclosed tessellation cells (ETCs). Serving as a proxy for the calculation of plot metrics, ETCs were generated from buildings, with streets, waterlines, and waterbodies specified as barriers following Arribas-Bel and Fleischmann (2022).

We compute 107 primary morphometric attributes providing a comprehensive urban form description, while avoiding a priori assumptions regarding metric relevance for LCZ discrimination. The set covers the majority of attributes available in *momepy* and covers five properties (dimension, shape, spatial distribution, intensity, and connectivity) across multiple spatial scales, ranging from the element itself, its immediate neighbors to larger neighborhoods defined by topological or metric distance. Additionally, we derive area-weighted variants for selected building and ETC shape and dimension characters to reflect the dominant local built environment and minimize the influence of smaller insignificant objects. A complete list of computed metrics is provided in the Supplementary Material A.

Subsequently, we derive contextual attributes to capture spatial patterns and smooth out local variation, hypothesizing their superior suitability for LCZ mapping operating at scales of hundreds of meters to several kilometers (Stewart and Oke, 2012). Prior to calculation, we link primary attributes to ETCs and for each, we compute the 25th, 50th, and 75th percentiles across neighboring ETCs within three topological steps of contiguity. These capture the local distribution, representing both the typical and extreme values of the primary attribute within its spatial context.

The final urban form description consists of 321 (107×3) contextual 2D morphometric attributes linked to individual ETCs, with the primary attributes being excluded from the final dataset.

## 3.2. Classifiers

To predict LCZ types from the morphometrics, either independently or coupled with satellite imagery, we employ two classifiers.

As the first classifier, we select RF due to its proven effectiveness in RS, particularly for handling multi-source, high-dimensional, and highly correlated data (Belgiu and Drăguţ, 2016). Compared to other classifiers, RF is less sensitive to overfitting and the quality of training samples, and it enables a selection of optimal input features through internal feature importance measurement (Belgiu and Drăguţ, 2016). We use the RF implementation provided by the *scikit-learn* Python library (Pedregosa et al., 2011).

As the second classifier, we use the Multi-Scale, Multi-Level Attention CNN MSMLA-50, which is a scene classification architecture based on SE-ResNet50 backbone, modified by integrated multi-scale and multi-level attention modules for the generation of spatially and spectrally enhanced features at multiple scales and hierarchical levels (Kim et al., 2021). The model was designed for LCZ mapping and has demonstrated superior performance compared to other CNN-based LCZ classifiers; for comparative evaluation and detailed architecture description, see original publication (Kim et al., 2021). For the purposes of this study, we translate the model to *PyTorch* (Paszke et al., 2019) and initialize the model with parameter filter depths set to 16, 32, 48.

## 3.3. Classification schemes

To implement and assess the proposed predictions, we design a framework of four classification schemes (S1–S4) (Figure 3). S1 predicts LCZ types based solely on morphometrics, thereby evaluating their predictive potential for LCZ mapping, whilst identifying a subset of key morphometrics for LCZ delineation, later used in fusion-based schemes (S3-S4). S2 represents the prevailing RS-based CNN approach, serving as a comparative baseline for the performance of the proposed methods, while also providing a pre-trained CNN for S4. Finally, S3 and S4 implement distinct techniques for fusing the identified morphometric subset with imagery, assessing whether this integration improves prediction performance relative to the RS-based baseline.

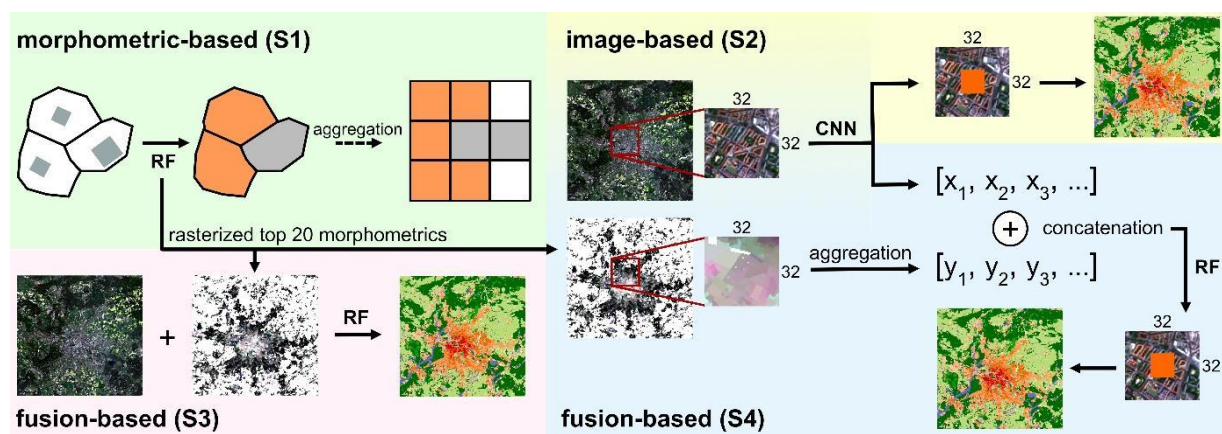

*Figure 3. Illustration of designed classification schemes.*

Scheme 1 (S1) implements the morphometric-based prediction, utilizing individual ETCs as spatial units and classifying them via RF based on the set of 321 contextual morphometric attributes. Notably, this approach is restricted to predicting urban LCZ types (LCZs 1–10), as natural types (LCZs A–G) inherently lack the built elements required for morphometric assessment. While we primarily evaluate the method's performance at the native ETC-level,

positing that LCZ mapping need not be confined to rigid grids, we also aggregate the output to 100 m grid (based on the majority LCZ type) solely to facilitate direct comparison with the other schemes. This scheme also identifies the subset of 20 most important attributes for LCZ delineation, derived from the Gini importance scores of the best-performing RF model.

Scheme 2 (S2) represents the RS-based CNN approach, here used as a baseline. This scheme utilizes the scene-based MSMLA-50 with the sliding window technique (Qiu et al., 2020). Satellite image patches of 320×320 m are generated with a 100 m step and fed into the trained CNN, resulting in a single LCZ label per patch. This label is assigned to the central 100×100 m area, producing a final LCZ map with a spatial resolution of 100 m.

Scheme 3 (S3) and Scheme 4 (S4) implement fusion-based LCZ prediction, integrating morphometrics with satellite imagery. Building on S1, they utilize the 20 most important attributes, rasterized to align with the spatial resolution of the imagery.

In S3, the rasterized subset is stacked with the satellite image as additional bands. Following the methodology of Bechtel et al. (2015) and Danylo et al. (2016), this raster is aggregated to a 100 m resolution using zonal statistics (mean, maximum, and minimum) and classified using RF, yielding a final LCZ map with a spatial resolution of 100 m.

In S4, the pre-trained MSMLA-50 from S2 is used in combination with the rasterized attributes and RF. Adopting the same technique as in S2, the 320×320 m image patches are passed through the CNN up to the layer preceding the classification head, yielding an aggregated multi-level attention feature vector. To incorporate the morphometric data, the rasterized attributes are aggregated from pixel-level to patch-level using five statistical functions (mean, minimum, maximum, standard deviation, and median). The resulting 100 (20×5) morphometric descriptors are concatenated with the CNN feature vector. This fused vector, describing the patch both spectrally and morphometrically, is subsequently classified using RF, ultimately creating LCZ map with a spatial resolution of 100 m.

### 3.4. Modelling and accuracy assessment

We perform implementation and modelling of all schemes separately for each study site, which is standard practice in LCZ mapping typically yielding the best results (Demuzere et al., 2019) as it naturally deals with spatial heterogeneity. Reference samples are split using stratified 5-fold cross-validation, with stratification based on ETC count for S1 and sample area for S2–S4, reflecting the difference in spatial units between ETC-based and image-based schemes.

The performance of individual models is assessed using weighted F1-score (F1), which is the average of class-wise F1-scores weighted by class sample counts, and overall accuracy (OA). Additionally, class-specific F1-scores were aggregated to evaluate performance for urban (F1U) and natural (F1N) types. The reported scheme metrics represent the average across the five models, while the confusion matrices are cumulative.

We fine-tune RF models by optimizing tree depth and the number of features considered at split to maximize OA on the test fold while avoiding overfitting. CNN training used probabilistic undersampling of majority classes and oversampling of minority classes together with data augmentation to balance classes and increase sample diversity. As with the RF models, we select the best-performing non-overfitted CNN model across the training epochs using model checkpointing. Details about the training are described in Supplementary Material B.

# 4. Data
## 4.1. LCZ reference data and study sites

The LCZ reference data are available from the 2017 IEEE GRSS Data Fusion Contest organized by the IEEE GRSS Image Analysis and Data Fusion Technical Committee in collaboration with WUDAPT and GeoWiki, with all samples revised by experts to ensure high accuracy (Yokoya et al. 2018; Tuia et al. 2019).

The dataset provides ground-truth polygon samples for five cities across three continents, namely Berlin, Hong Kong, Paris, Rome, and São Paulo, which we use as study sites (Figure 4). These cities have diverse geographic, cultural, and historical backgrounds, reflected in their urban structure, making them a robust set of sites for this study.

However, the reference datasets are highly imbalanced. Not all LCZ types are present in every city, and the number and size of polygon samples vary between types and cities. LCZ-7 (Lightweight low-rise) is absent from all sites because it is not present in their urban fabric. In two cases, a type is represented by a single polygon, LCZ-10 (Heavy industry) in Rome and LCZ-9 (Sparsely built) in Paris. We manually split these into two parts to ensure that the type always appears in the training folds during the adopted cross-validation.

The quantitative characteristics of the reference data, comprising polygon and ETCs counts, and area coverage across all sites, are detailed in the Supplementary Material C.

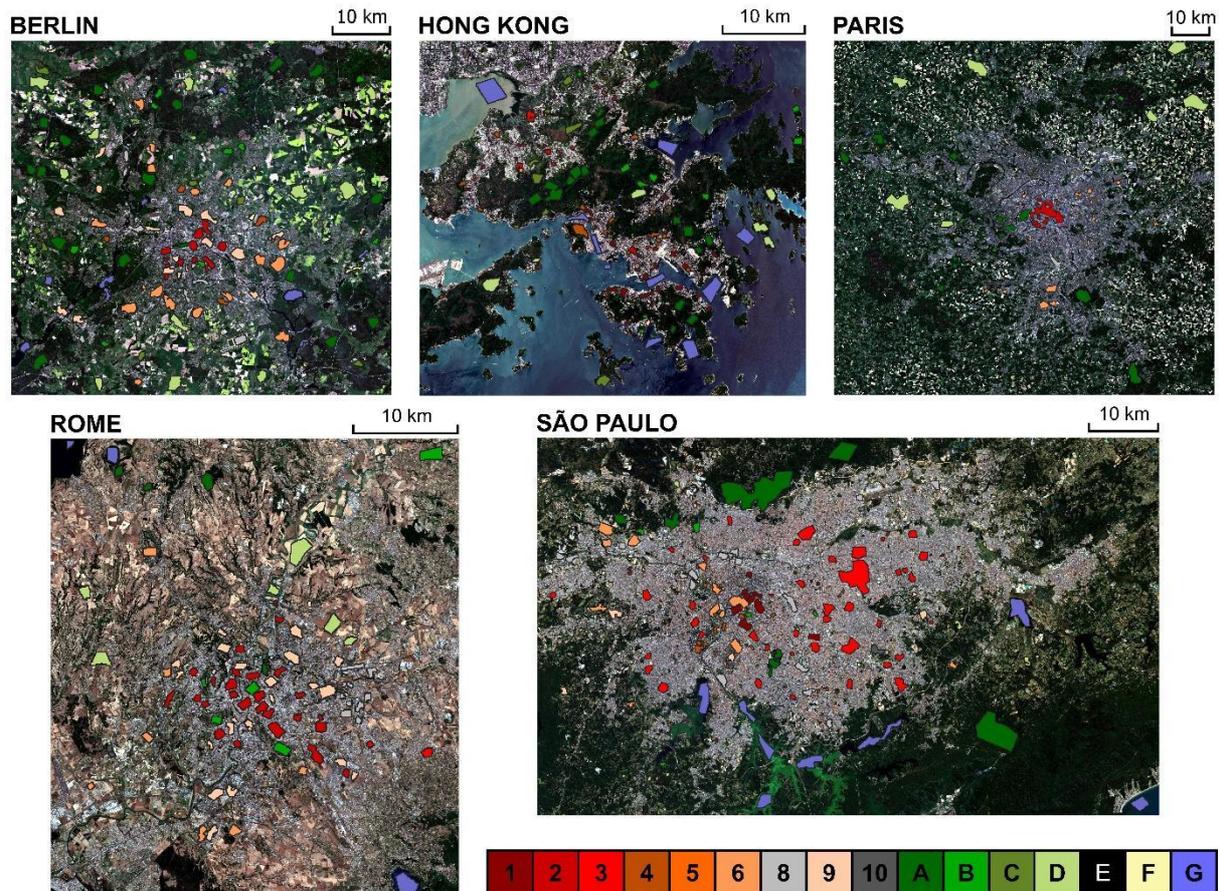

Figure 4. Study sites, LCZ reference data, and Sentinel-2 imagery.

## 4.2. Morphological data

For each site, we retrieved building footprints, street networks, waterlines and waterbodies from the 2025-02-19.0 release of Overture Maps. However, this causes a temporal discrepancy between these datasets (2025) and the reference data (2017) due to the unavailability of Overture Maps data for the earlier year. This may introduce inconsistency between the morphometric input and the references, potentially affecting prediction accuracy.

Prior to morphometric analysis, building footprints and street networks were preprocessed to address quality issues and ensure geometric and topological integrity as outlined in the Supplementary Material D.

## 4.3. Satellite imagery

We use cloud-free, atmospherically corrected Sentinel-2 multispectral Level-2A scenes taken on a single date during the growing season from April to September of 2017 from the Copernicus Data Space Ecosystem. For Hong Kong, we use a late March 2018 scene due to the lack of cloud-free imagery in 2017. The specific Sentinel-2 products used are listed in the Supplementary Material E.

We use ten spectral bands: the 10 m bands (Blue, Green, Red, NIR) and the 20 m bands (four Red Edge and two SWIR), which we resampled to 10 m using bilinear resampling.

# 5. Results
## 5.1. Morphometric-based prediction

The overall performance of the morphometric-based LCZ prediction (S1), classifying ETCs into urban LCZ types (LCZs 1–10) based solely on morphometrics and evaluated at the native ETC-level, varies substantially across sites, with F1 ranging from 64.2% for Rome to 92.5% for Paris (Figure 5). Furthermore, we observe pronounced differences between the best and worst-performing folds ranging from 13% to over 57% for the F1, indicating model instability and high sensitivity to the specific composition of the training data.

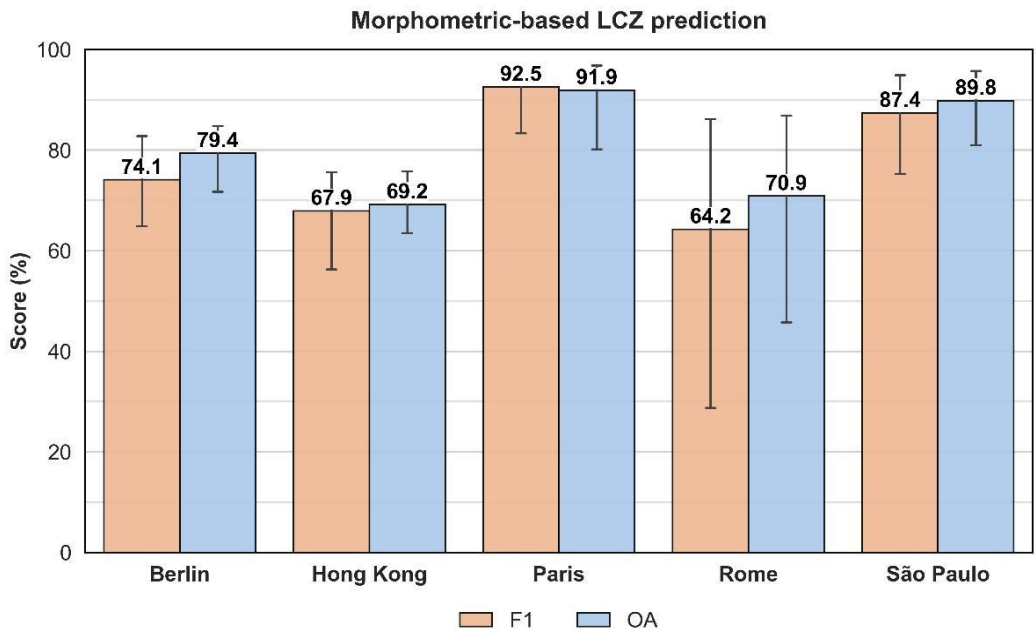

*Figure 5. Overall performance of the morphometric-based LCZ prediction, evaluated at the ETC-level.*

The confusion matrices (Figure 6) indicate mixed success in differentiating types varying in building density. Compact types (LCZs 1–3) are generally well distinguished from open arrangement types (LCZs 4–6), and vice versa, with more pronounced misclassifications largely confined to less common types, except in Rome, where this distinction is very weak. In contrast, the sparsely built type (LCZ-9) is systematically misclassified as the more prevalent open low-rise (LCZ-6) across all sites where it occurs, reflecting difficulty in distinguishing sparse from open arrangements.

Distinguishing types differing by building height (among LCZs 1–3, or LCZs 4–6) is limited, with successful differentiation varying by site. Generally, accurate identification remains moderate and is largely confined to the predominant height stratum of each site, while less common height categories are frequently misclassified as the dominant height type.

Large low-rise areas (LCZ-8) are distinguished from open low-rise (LCZ-6) with only minor confusion, confirming building size discernibility within open low-rise arrangements, but suffer mainly from confusion with other taller open types, further supporting previous findings.

The results for Heavy industry (LCZ-10) highlight the difficulty in delineating industrial areas, evidenced by its complete misclassification in Rome and only marginal detection in São Paulo. In contrast, in Hong Kong a substantial share of samples is correctly identified, although some other types are misclassified as LCZ-10.

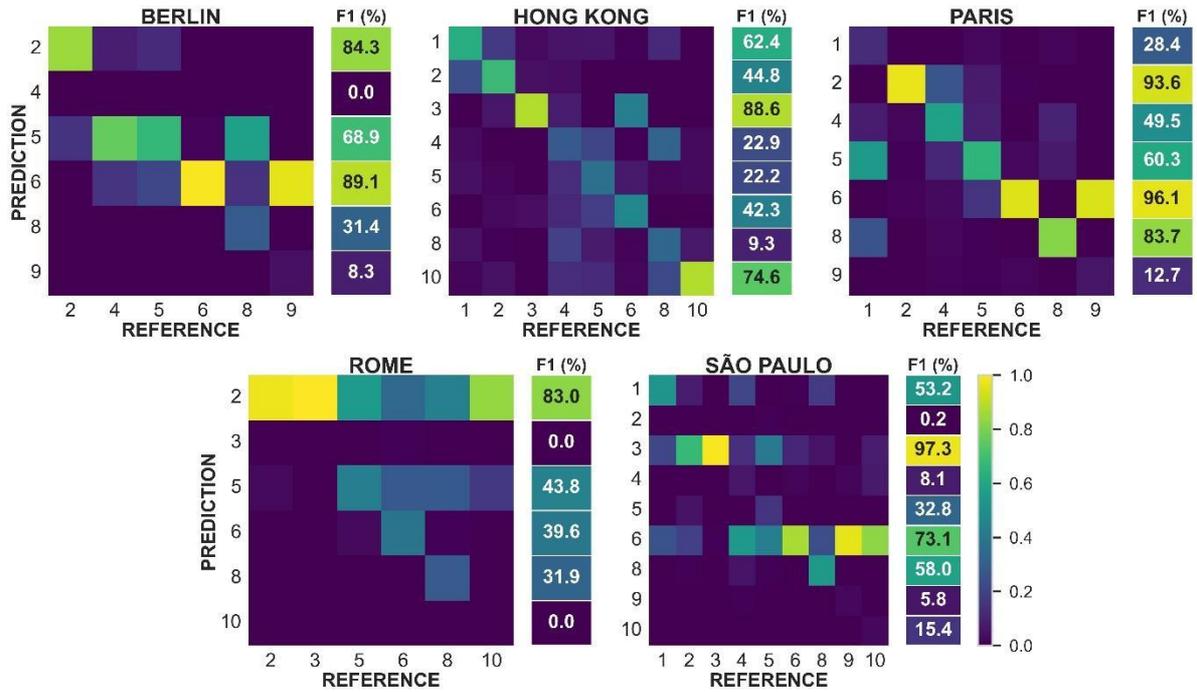

Figure 6. The confusion matrices of the morphometric-based LCZ prediction with average class F1-scores, evaluated at the ETC-level.

## 5.2. Fusion-based prediction and comparisons with RS-based method

The RS-based method (S2) served as the baseline for evaluating the morphometric-based prediction (S1), in this case aggregated from ETC-level to 100 m grid, and the fusion-based predictions (S3–S4), integrating Sentinel-2 spectral data with the site's 20 most important morphometric attributes identified in S1. The specific morphometric subsets are provided in the Supplementary Material F.

When comparing the aggregated S1 against other methods in terms of F1U (Figure 7), S1 performed the worst by a significant margin across four sites, with exception of Paris, where it slightly outperformed both the RS-based and fusion-based methods (+1.5% relative to S2 and +0.5% relative to S3). Furthermore, the discussed model instability and sensitivity to training data are most pronounced in S1 compared to other methods.

The comparison between RS-based and fusion-based predictions shows that the impact of the fusion is generally modest and inconsistent (Figure 7). Considering each site's better-performing fusion-based scheme (S3 or S4), determined by combined F1 and F1U to prioritize urban types performance alongside the overall performance, the F1U improved relative to S2 in Berlin (1.3%, S3), Hong Kong (4%, S3), Paris (0.9%, S4), and Rome (6.4%, S3), but declined in São Paulo (1.7%, S4). In terms of F1N, a decrease of 0.2% to over 10% is observed, except for a slight increase of 1.3% in Rome. As a result, the F1 improved only in Paris (0.1%, S4) and Rome (1%, S3), while for other sites it decreased by 2 – 3%. Moreover, the model instability propagated also to the fusion schemes and is more acute compared to S2.

Based on the previous combined assessment, neither fusion scheme proves universally superior to the other. However, in terms of F1U, S3 clearly outperformed S4 at three out of five sites with gains ranging from approximately 4% to over 8%, while performing comparably in the remaining two. Conversely, in terms of F1N, S4 appears more effective, showing gains of

3% to over 12% in three sites, while performing slightly worse in the others. This suggests that S3 is better suited for urban LCZ types, whereas S4 for natural LCZ types.

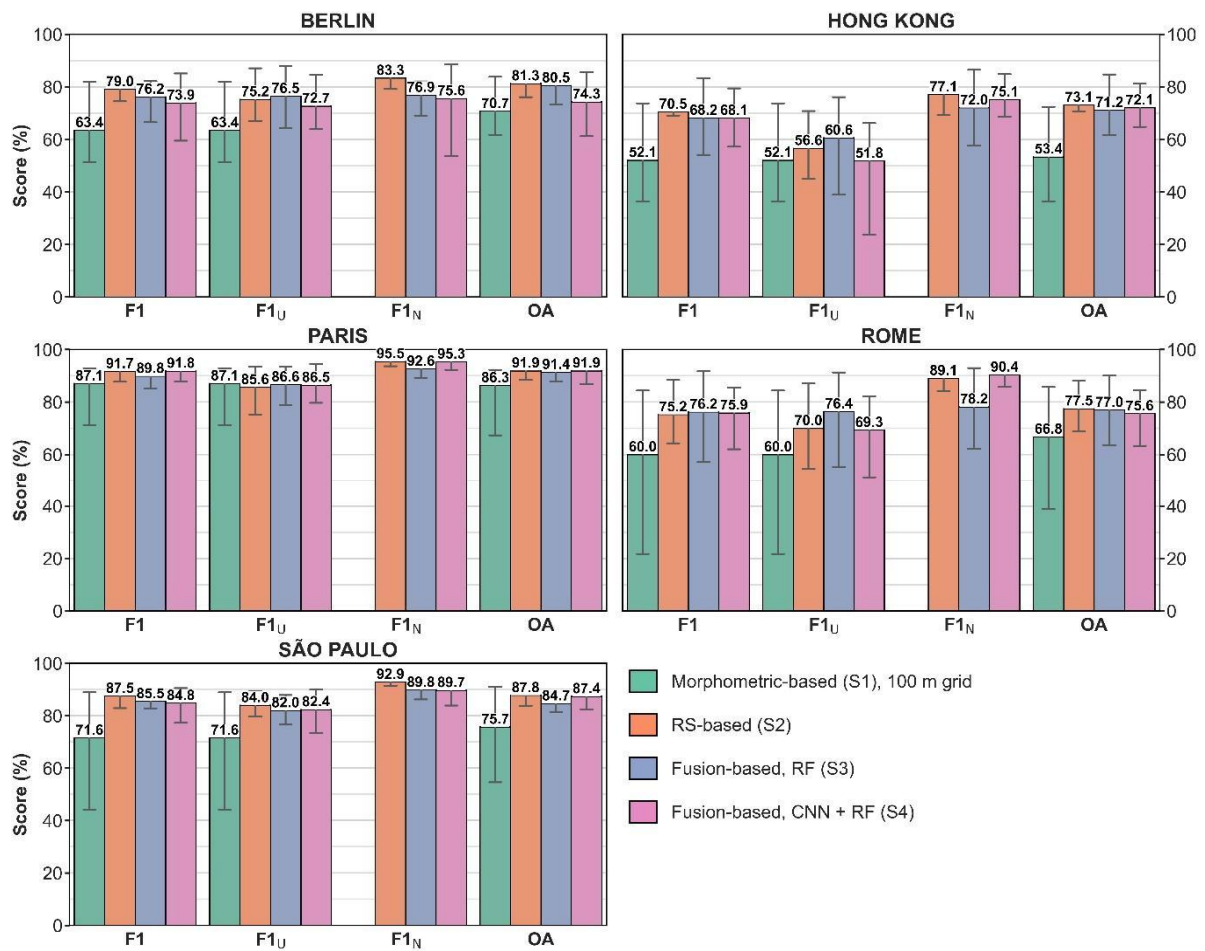

*Figure 7. Comparison of the morphometric-based prediction, aggregated to 100 m grid, and fusion-based predictions with the RS-based method.*

A comparison of the confusion matrices between S2 and the best-performing fusion scheme (Figure 8) reveals that in Hong Kong (S3) and Rome (S3), spectral and morphometric data complemented each other effectively, improving the delineation for the majority of types, although misclassification among certain types increased, particularly those distinguished by building height. In contrast, the effect of fusion in Berlin (S3) and Paris (S4) was generally minimal; with the performance of several types remaining essentially unchanged, while others showed only minor fluctuations (improvements or declines). In São Paulo (S4), the majority of types exhibited a decline, characterized by worsened discrimination of building density, height, and industrial function, while the remaining types largely remained unchanged.

Regarding the systematic misclassification of sparsely built (LCZ-9) as open low-rise (LCZ-6) observed in S1, such misclassification appears also in the RS-based S2 prediction, although significantly reduced. In this context, the combination of spectral and morphometric data proved detrimental across sites, resulting in increased misclassification compared to S2.

Conversely, for Heavy industry (LCZ-10), accuracy substantially improved across all sites under the S3 fusion relative to the baseline. Under S4, however, performance declined everywhere with the exception of a slight increase in Hong Kong, suggesting that S3 might be more suitable method for predicting this type. Notably, LCZ-10 is the only type to exhibit such pattern.

Regarding natural LCZ types, confusion patterns in Rome (S3) and Paris (S4) showed minimal change relative to S2. In the remaining three cities, some natural types remained unaffected, most consistently dense trees (LCZ-A) and water (LCZ-G), but declines were more frequent and often pronounced, particularly evident for scattered trees (LCZ-B), which was increasingly confused with dense trees or low plants (LCZ-D). Furthermore, confusion of natural types with urban types was not resolved by the fusion and for certain classes it actually increased.

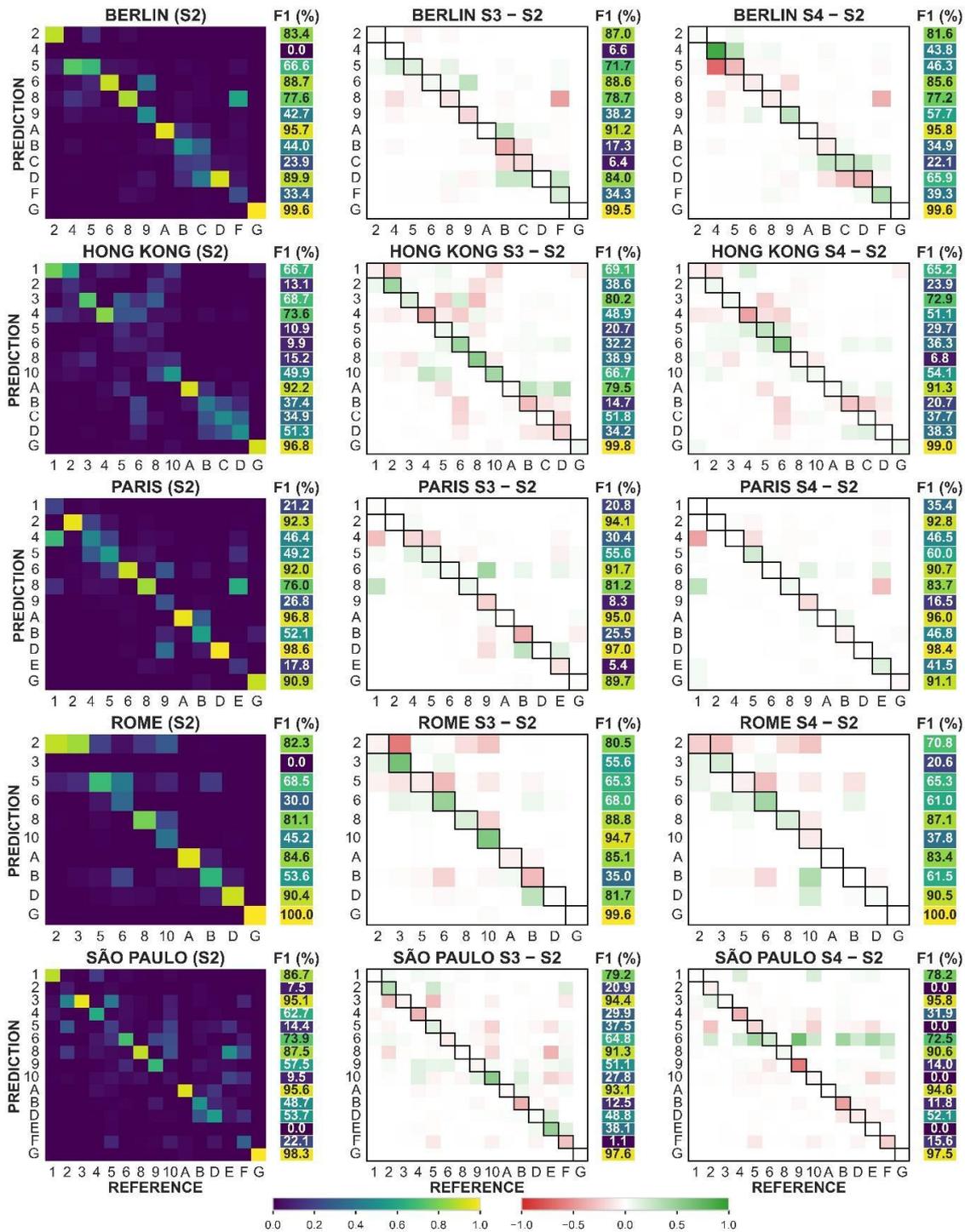

Figure 8. Comparison between the confusion matrices and average class F1-scores of RS-based and fusion-based predictions.

## 5.3. LCZ Maps

Figure 9 displays the LCZ maps generated by the best-performing fold of morphometric-based (at native ETC-level), RS-based, and the better-performing fusion-based scheme (S3 or S4) across the five sites.

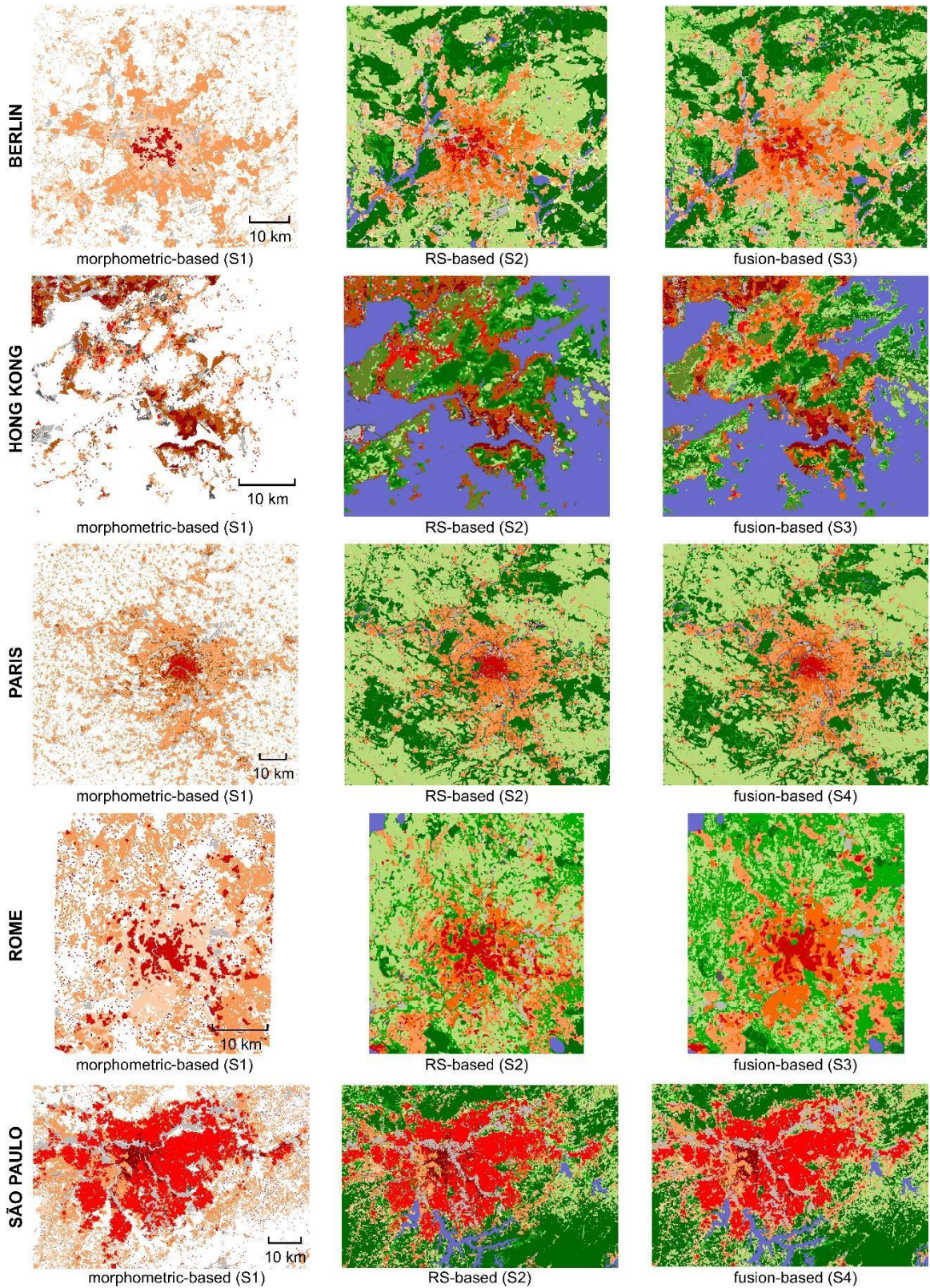

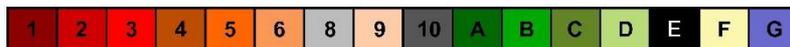

*Figure 9: LCZ maps of the best-performing fold of morphometric-based, RS-based, and the better-performing fusion-based scheme.*

# 6. Discussion & Conclusion

The results of morphometric-based LCZ prediction indicate that the correspondence between 2D urban morphometrics and urban LCZ types is selective, ultimately making the efficacy of this method site-dependent. While the delineation of compact versus open types, as well as varying building sizes within open low-rise arrangements, can be accurate, the discrimination between sparse and open arrangements, types differing by building height, and industrial areas remains limited. These limitations fundamentally constrain the achievable accuracy of this method. Based on the performance observed across our study sites, we conclude that the predictive potential of 2D urban morphometrics is high in sites dominated by two LCZ types differing in compact versus open building arrangement, as well as in sites characterized by the coexistence of large low-rise structures (LCZ-8) and open low-rise types (LCZ-6). Conversely, this potential diminishes in the presence of building height variability, industrial areas (LCZ-10), or the coexistence of sparsely built (LCZ-9) and open low-rise (LCZ-6) areas.

Moreover, the relationship between 2D urban morphometrics and LCZ typology also appears to be inconsistent, as evidenced by the pronounced model instability and sensitivity to the specific composition of training data, observed not only in the morphometric-based but also in the fusion-based predictions, and to a degree significantly higher than in the RS-based approach.

However, compared to previous studies, our application of a comprehensive morphometric urban form characterization for LCZ prediction demonstrated that a much broader range of urban form properties are relevant for distinguishing urban LCZ types. Based on the key morphometrics identified in S1, we confirm the significance of established building spatial distribution metrics (e.g., interbuilding distance) and intensity metrics (e.g., coverage area ratio). Crucially, our results also highlight the relevance of building shape metrics (e.g., facade ratio, fractal dimension) and complex dimension metrics (e.g., perimeter wall length, longest axis length), along with descriptors of the street network (e.g., street openness, mean node degree) and plots (e.g., dimensions and shape). Therefore, future LCZ mapping methods integrating numerical urban form descriptions could consider these metrics alongside established parameters.

The fusion of morphometrics and imagery aimed to improve the accuracy of urban LCZ type delineation over the prevailing RS-based method. While relatively distinct improvements were observed in Hong Kong and Rome (ranging from 4% to over 6%), gains at two other sites were practically negligible (around 1%), and the final site even exhibited a slight decline. Notably, in Hong Kong and Rome, where both the RS-based and the morphometric-based prediction struggled, the improvement achieved by fusion highlighted the potential of effective combination of spectral and morphometric data. Therefore, 2D urban morphometrics combined with spectral data can enhance urban LCZ mapping performance relative to the RS-based method, albeit inconsistently and perhaps not significantly.

Among previous studies, only Yoo et al. (2020) and Ao et al. (2025) compared the RS-based method with an approach combining imagery and urban form metrics for LCZ prediction. However, beyond relying on a single urban form metric, these studies also have other limitations. While Ao et al. (2025) reported an improvement of over 10% for every urban LCZ type accuracy using a fusion of imagery and building density, their evaluation was restricted to only two study sites, which limits the robustness of their conclusions. Moreover, they employed pixel-based Random Forest (RF) prediction for both approaches, a method that achieves lower accuracy than scene-based CNN classification (Rosentreter et al. 2020), thus likely disadvantaging the RS-based prediction. Yoo et al. (2020) employed scene-based CNN for both the RS baseline and the fusion of imagery with building density and, unlike our study, building height data. However, their testing was similarly limited to just two sites. Comparing the approaches, they observed inconsistent outcomes, with one site improving by over 7% and the other declining by 1%, which aligns with our results. Crucially, both studies reported performance based on a single model per site, limiting the robustness of their findings. In contrast, we evaluated each approach across five diverse sites, derived robust performance estimates by averaging results across five distinct models, and utilized scene-based CNN as the RS baseline. Furthermore, we designed and tested two distinct fusion methods to explore the potential of combining spectral and morphometric data more thoroughly, rather than relying on a single technique.

A limitation of our study is the omission of building height data; however, we argue that 2D urban morphometrics can serve as a sufficient proxy for building height (Milojevic-Dupont et al., 2020). Another limitation concerns the reference data, specifically the small amount of samples per site, which for this particular reference dataset also confirms Yoo et al. (2019). For some LCZ types across the sites, the number of samples was below five, meaning that not all folds in the adopted 5-fold cross-validation contained examples of every LCZ type present at a given site, which may have affected the stability of reported performance metrics. Furthermore, limited amount of samples can negatively impact class-specific accuracy; this factor likely contributed especially to the poor performance of Heavy industry (LCZ-10).

Future research could focus on identifying a parsimonious, transferable subset of morphometric attributes with the highest relevance for LCZ prediction. Further investigation is also required to determine the specific conditions under which combining spectral and morphometric data enhances performance, and to which extent, ultimately clarifying when fusion offers an advantage over RS-based method. Closely linked to this is the development of a new fusion technique, as none of the techniques evaluated here demonstrated consistent superiority. Finally, future studies could explore the synergistic integration of 2D urban morphometrics with 3D building metrics derived from 3D city models (Labetski et al., 2023), representing a logical extension of our study and another novelty in LCZ mapping.

The selective and inconsistent relationship observed between 2D urban morphometrics and the urban LCZs indicates that the correspondence between LCZ typology and the measurable aspects of urban form, including the spatial relationships of its elements, is generally tenuous. This highlights the conceptual limitations of the LCZ typology regarding urban form description, ultimately restricting its interpretability and analytical value in a morphological context. Building on these findings, we argue that LCZ framework should not be considered a substitute for morphological urban form analysis, as is common practice (Taubenböck et al., 2020; Naserikia et al., 2023; Debray et al., 2025), and should be used with caution in morphological studies, as it risks oversimplification, internal inconsistency, reduced comparability across sites, thereby compromising our understanding of how urban structure shapes, and is shaped by, environmental and socioeconomic performance.

# 7. Code & data availability

Code, with the reproducible environment, is available at github.com/majerhugo/lcz_morphometrics and archived at doi.org/10.5281/zenodo.18684979. Data used are available at doi.org/10.5281/zenodo.18692973, and original LCZ reference data are available at dx.doi.org/10.21227/e56j-eh82.

# 8. Acknowledgements

The authors kindly acknowledge funding by the Charles University's Primus programme through the project "Influence of Socioeconomic and Cultural Factors on Urban Structure in Central Europe", project reference PRIMUS/24/SCI/023. We would also like to thank WUDAPT, the IEEE GRSS Image Analysis and Data Fusion Technical Committee, and the contributors for LCZ reference samples, in particular Chao Ren, Daniel Fenner, Dragan Milosevic, Guillaume Dumas, and Maria De Fatima Andrade.

# Supplementary Material A: Urban Morphometrics calculation details

Prior to the calculation of morphometrics, we linked ETCs to their respective parent building, nearest street, node and edge and excluded those without parent building. We compute the morphometrics at multiple spatial scales defined as:

- Neighborhood within distance 20 m, 100 m or 200 m (used for building-based metrics),
- 10, 20 or 30 nearest neighbors (used for building-based metrics),
- 1, 2 or 3 topological steps (used for ETC-based metrics),
- 5 m or 400 m radii (used for street-based metrics).

Each metric was calculated using the *momepy* Python library (Fleischmann, 2019) and its tools. The complete list of computed primary morphometrics (below) includes all parameters used for their computation. For metric's details and implementation see the linked *momepy* documentation.

**1) Dimensional & Shape Morphometrics**

We computed each of the dimensional and shape attributes for [Buildings] and [ETCs], unless specified. The *weighted versions* correspond to weighting by area of the objects (buildings or ETCs) within specified neighborhood [doc link].

- Area: (*No weighted versions*)
- Courtyard Area [doc link]: (Buildings only). (*No weighted versions*)
- Courtyard Index [doc link]: (Buildings only). (*No weighted versions*)
- Perimeter Wall Length of joined buildings [doc link]: (Buildings only). +*Weighted versions: Buildings (100 m, 200 m).*
- Longest Axis Length [doc link]: +*Weighted versions: Buildings (100 m, 200 m), ETCs (3 topological steps)*
- Circular Compactness [doc link]: +*Weighted versions: Buildings (100 m, 200 m), ETCs (3 topological steps).*
- Square Compactness [doc link]: +*Weighted versions: Buildings (100 m, 200 m), ETCs (3 topological steps).*
- Compactness-weighted axis [doc link]: +*Weighted versions: Buildings (100 m, 200 m), ETCs (3 topological steps).*
- Convexity [doc link]: +*Weighted versions: Buildings (100 m, 200 m), ETCs (3 topological steps).*
- Elongation [doc link]: +*Weighted versions: Buildings (100 m, 200 m), ETCs (3 topological steps).*
- Equivalent Rectangular Index [doc link]: +*Weighted versions: Buildings (100 m, 200 m), ETCs (3 topological steps).*
- Facade Ratio [doc link]: +*Weighted versions: Buildings (100 m, 200 m), ETCs (3 topological steps).*
- Fractal Dimension [doc link]: +*Weighted versions: Buildings (100 m, 200 m), ETCs (3 topological steps).*
- Rectangularity [doc link]: +*Weighted versions: Buildings (100 m, 200 m), ETCs (3 topological steps).*
- Shape Index [doc link]: +*Weighted versions: Buildings (100 m, 200 m), ETCs (3 topological steps).*
- Square Compactness [doc link]: +*Weighted versions: Buildings (100 m, 200 m), ETCs (3 topological steps).*

**2) Spatial Distribution & Intensity Morphometrics**

- Building Adjacency: The level of building adjacency measured within 200 m neighborhood [doc link]
- Mean interbuilding distance: Mean distance between buildings measured within 200 m neighborhood [doc link]
- Shared Walls: The length of shared walls of adjacent buildings [doc link]
- Street Alignment: The deviation of the building orientation from the street orientation. [doc link] +*Weighted versions: Building (100 m, 200 m)*
- Coverage Area Ratio: Proportion of an ETC covered by a building. +*Weighted version: ETC (3 topological steps)*
- ETC Granularity: Sum of ETCs area measured within 1 topological step. [doc link]
- Neighbors (number of neighbors) [doc link]:
  o Buildings: measured within 20 m, 100 m, and 200 m neighborhood.
  o ETCs: measured within 1, 2, and 3 topological steps.
- Mean Distance to Neighbors [doc link]:
  o Buildings: measured within 20 m, 100 m, and 200 m neighborhood; and for 10, 20, and 30 nearest neighbors.
  o ETCs: measured within 2, and 3 topological steps.

**3) Street Descriptors & Connectivity Morphometrics**

- Street Length
- Street Linearity [doc link]
- Width, Width Deviation, Openness [doc link]
- Node Degree for each node [doc link]
- Mean Node Degree within 5 m and 400 m radii [doc link]
- Mean Node Distance: mean distance to neighboring nodes [doc link]
- Node Density: the density of a node's neighbors within 5 m and 400 m radii [doc link]
- Clustering: the squares clustering coefficient for nodes [doc link]
- Edge to Node Ratio: the number of edges to the number of nodes within 5 m and 400 m radii [doc link]
- Length of Cul-De-Sacs within 5 m and 400 m radii [doc link]
- Cyclomatic complexity within 5 m and 400 m radii [doc link]
- Connectivity gamma index within 5 m and 400 m radii [doc link]
- Meshedness within 5 m and 400 m radii [doc link]

This yielded 107 primary morphometrics, which we later joined to ETCs geometries. Subsequently, we computed their 25th, 50th, and 75th percentiles across neighboring ETCs within three topological steps [doc link], creating a set of 321 morphometric attributes.

## Supplementary Material B: Modelling details

For the training of the RF models (used in the morphometric-based S1, and fusion-based predictions S3-S4), we independently applied and evaluated two weighting strategies across all five folds of each scheme: (1) equal weights for all classes; (2) class weights inversely proportional to class frequencies in the input data.

In S1, this may yield two different subsets of the 20 most important morphometric attributes for a site: 1) one derived from the best-performing RF model without weighting; 2) the other from the best-performing RF model using class weighting. To assess which subset performs better in the fusion-based schemes (S3-S4), both subsets and both weighting strategies of S3 and S4 RF models were tested and evaluated separately across their respective folds (resulting in four possible configurations for each scheme). Ultimately, we report only S3 and S4 results based on the better-performing subset of morphometrics. Additionally, presented results of all schemes that utilize the RF classifier correspond to the better-performing weighting strategy. In both cases, we determined the better result by comparing combined F1 and F1U values (averaged across the folds).

We fine-tuned each RF model individually by adjusting the maximum tree depth and the number of features considered at each split, as tree depth plays a key role in model generalization and together with the number of features considered at each split has the strongest influence on classification accuracy (Contreras et al., 2021). We selected their combination based on the highest OA achieved on the test fold. To avoid overfitting, we considered only models with a generalization gap, defined as the difference between OA on the training and test set, smaller than 5%. If no parameter combination satisfied this criterion, we selected the model with the smallest generalization gap.

The training of the CNN models in S2 utilized oversampling of minority classes and undersampling of majority classes via PyTorch's probabilistic-based sampling method *WeightedRandomSampler*. Firstly, for each site, a reference set of non-overlapping 320×320 m image patches with central class-defining pixel located within a reference polygon was created using the described sliding window approach (Qiu et al., 2020). Based on the fold assignment of the corresponding reference polygon, each patch was included either in the training or testing set during the 5-fold cross-validation procedure. Individual samples weights, here representing the sampling probabilities, were based on the class of a sample and set as inverse class frequencies smoothed with exponent $\alpha=0.5$, mainly to prevent excessive undersampling of majority classes. Similarly, the number of samples drawn per epoch was set to twice the size of the original training set to increase the probability of selecting a larger portion of samples from majority classes, while also ensuring sufficient representation of minority classes. To increase the diversity of samples and to perform oversampling of minority classes, augmentation techniques random horizontal flip, vertical flip, and rotation were applied to all samples. Prior to training, we normalized all samples using the mean and standard deviation derived exclusively from the training set.

We trained each CNN model from scratch over 100 epochs, using a learning rate of 0.0001, the Adam optimizer, the CrossEntropy loss function, and a batch size of 128. As with the RF models, we selected the best-performing non-overfitted model across the epochs using model checkpointing, defined as the one achieving the highest OA on the test fold with a generalization gap not exceeding 5%.

# Supplementary Material C: Quantitative characteristics of LCZ reference data

Table C1 shows for each site and LCZ type the corresponding number of reference polygons (#), and the stratification variables for reference data split – ETCs counts and total reference area.

| LCZ | Berlin | | | Hong Kong | | | Paris | | | Rome | | | São Paulo | | |
| --- | --- | --- | --- | --- | --- | --- | --- | --- | --- | --- | --- | --- | --- | --- | --- |
| | # | ETCs | Area (ha) | # | ETCs | Area (ha) | # | ETCs | Area (ha) | # | ETCs | Area (ha) | # | ETCs | Area (ha) |
| 1 | - | - | - | 26 | 3460 | 631 | 2 | 105 | 56 | - | - | - | 12 | 17552 | 955 |
| 2 | 13 | 13525 | 1534 | 11 | 2983 | 179 | 11 | 43711 | 2705 | 25 | 11192 | 1551 | 12 | 2601 | 134 |
| 3 | - | - | - | 14 | 7924 | 326 | - | - | - | 2 | 1136 | 104 | 41 | 248129 | 5308 |
| 4 | 6 | 1555 | 577 | 19 | 1723 | 673 | 12 | 1758 | 366 | - | - | - | 14 | 3518 | 482 |
| 5 | 12 | 19155 | 2448 | 8 | 600 | 126 | 13 | 5520 | 446 | 23 | 4814 | 1495 | 8 | 4437 | 244 |
| 6 | 17 | 66963 | 4010 | 13 | 1217 | 120 | 32 | 60639 | 2419 | 7 | 2113 | 480 | 12 | 28013 | 1862 |
| 7 | - | - | - | - | - | - | - | - | - | - | - | - | - | - | - |
| 8 | 14 | 2967 | 1654 | 9 | 154 | 137 | 10 | 1409 | 748 | 11 | 681 | 435 | 18 | 9205 | 1915 |
| 9 | 7 | 11095 | 761 | - | - | - | 2 | 315 | 59 | - | - | - | 3 | 2601 | 335 |
| 10 | - | - | - | 9 | 822 | 219 | - | - | - | 2 | 61 | 49 | 2 | 943 | 179 |
| A | 18 | - | 4960 | 14 | - | 1616 | 12 | - | 4497 | 5 | - | 284 | 8 | - | 6359 |
| B | 9 | - | 1028 | 15 | - | 540 | 6 | - | 394 | 5 | - | 555 | 6 | - | 302 |
| C | 7 | - | 1050 | 13 | - | 691 | - | - | - | - | - | - | - | - | - |
| D | 12 | - | 4424 | 16 | - | 985 | 8 | - | 7688 | 7 | - | 984 | 12 | - | 376 |
| E | - | - | - | - | - | - | 7 | - | 214 | - | - | - | 5 | - | 109 |
| F | 6 | - | 359 | - | - | - | - | - | - | - | - | - | 10 | - | 144 |
| G | 10 | - | 1732 | 10 | - | 2603 | 6 | - | 234 | 3 | - | 500 | 8 | - | 3492 |

Table C1. Number of reference polygons (#), ETCs, and the total reference area per LCZ type on each site.

# Supplementary Material D: Morphological data

The retrieved building footprints correspond to the type *building* in the Overture Maps data scheme. Only the geometry of building footprints was considered, all additional attributes were removed.

The street network datasets consist of features of type *segment* and subtype *road*. Road features of class *service* were omitted.

For the water datasets, the corresponding type is *water*. The linestring features of this type form the waterlines dataset, the polygon features the waterbodies dataset. We filtered out underground and aboveground water features. From waterbodies we excluded features of subtypes *human-made*, *physical*, *reservoir*, *spring* and *wastewater*.

*Preprocessing*

Prior to the urban morphometrics computation, we preprocessed the building footprint and street network dataset of each site to mitigate their quality issues, ensure their geometric and topological integrity and overall suitability for morphological analysis of urban form.

In particular, the preprocessing of building footprint dataset involved fixing invalid geometries, exploding multipolygon features, removing non-polygon features, excluding excessively large buildings, and simplification of the footprint shapes. To address topological issues such as overlapping buildings, we performed merging and trimming of affected features. Additionally, small buildings touching larger ones were merged into the adjacent larger structures. This workflow was carried out primarily using the *geoplanar* (Rey et al., 2025) and *GeoPandas* (Van den Bossche et al., 2025) Python libraries.

We preprocessed the street networks to create a closer representation of morphological network rather than transportational. Firstly, tunnels longer than 50 m were identified and removed, as such features have little impact on the morphological structure of a city. Subsequently, the networks were simplified using *neatnet* (Fleischmann et al., 2026) Python library. This automatized parameter-free simplification method mainly consists of collapsing dual carriageways to a single line, and replacement of roundabouts and complex intersections by a single node. We used the preprocessed building footprints as an exclusion mask parameter, which serves for identifying network parts that should be preserved without any change.

Lastly, we conducted a topological consistency check across datasets. Buildings intersecting with streets or waterbodies were removed, as were waterlines intersecting buildings. In the first case, the removal of individual buildings was deemed more appropriate than eliminating significant segments of main or local streets, in the second case intersecting features typically represented various industrial ancillary objects or structures. The removed waterline features primarily consisted of small local streams or drainage channels.

# Supplementary Material E: Satellite Imagery

Table E1 shows the specific Sentinel-2 products used.

| Berlin | S2B_MSIL2A_20170519T101019_N0500_R022_T32UQD_20231115T232025 |
|---|---|
| | S2B_MSIL2A_20170519T101019_N0500_R022_T33UUU_20231115T232025 |
| | S2B_MSIL2A_20170519T101019_N0500_R022_T33UVT_20231115T232025 |
| | S2B_MSIL2A_20170519T101019_N0500_R022_T33UVU_20231115T232025 |
| Hong Kong | S2A_MSIL2A_20180321T025541_N0500_R032_T49QGE_20230906T145622 |
| | S2A_MSIL2A_20180321T025541_N0500_R032_T49QHE_20230906T145622 |
| Paris | S2A_MSIL2A_20170526T105031_N0500_R051_T31UCP_20231113T160651 |
| | S2A_MSIL2A_20170526T105031_N0500_R051_T31UCQ_20231113T160651 |
| | S2A_MSIL2A_20170526T105031_N0500_R051_T31UDP_20231113T160651 |
| | S2A_MSIL2A_20170526T105031_N0500_R051_T31UDQ_20231113T160651 |
| Rome | S2A_MSIL2A_20170620T100031_N0500_R122_T32TQM_20231009T162856 |
| | S2A_MSIL2A_20170620T100031_N0500_R122_T33TUG_20231009T162856 |
| São Paulo | S2B_MSIL2A_20170726T131239_N0500_R138_T23KLP_20230904T114104 |
| | S2B_MSIL2A_20170726T131239_N0500_R138_T23KLQ_20230904T114104 |

*Table E1. Specific Sentinel-2 products used.*

Firstly, for each site, we united the coordinate reference systems of the tiles, and then we mosaicked them. All imagery preprocessing was performed in QGIS.

# Supplementary Material F: Subsets of morphometrics for fusion-based LCZ predictions

Figure F1 presents the subsets of the 20 most important morphometric attributes for each site, derived from the best-performing RF model across folds in S1, subsequently utilized in the fusion-based LCZ prediction schemes (S3-S4).

For Berlin, Hong Kong, and Rome, subsets from both unweighted and class-weighted RF models are displayed, as S3 yielded better results using the unweighted set, while S4 performed superiorly with the class-weighted set. For Paris, the fusion-based predictions performed better using the unweighted set, whereas for São Paulo, the class-weighted set proved superior.

The subsets differ across study sites, although the longest axis length of building, building facade ratio, fractal dimension of building, and coverage area ratio appear consistently across all sites. Four sites shared the compactness-weighted axis of building and street openness, while three sites shared the perimeter wall length of joined buildings, mean interbuilding distance and shape index, square compactness, and equivalent rectangular index of a building.

Characters derived from all considered morphological elements are included in the subsets, indicating that each element provides useful information for LCZ discrimination, although building-based metrics dominate across sites. In addition, the area-weighted versions of characters describing shape and dimension of building or ETC prevail compared to corresponding unweighted characters.

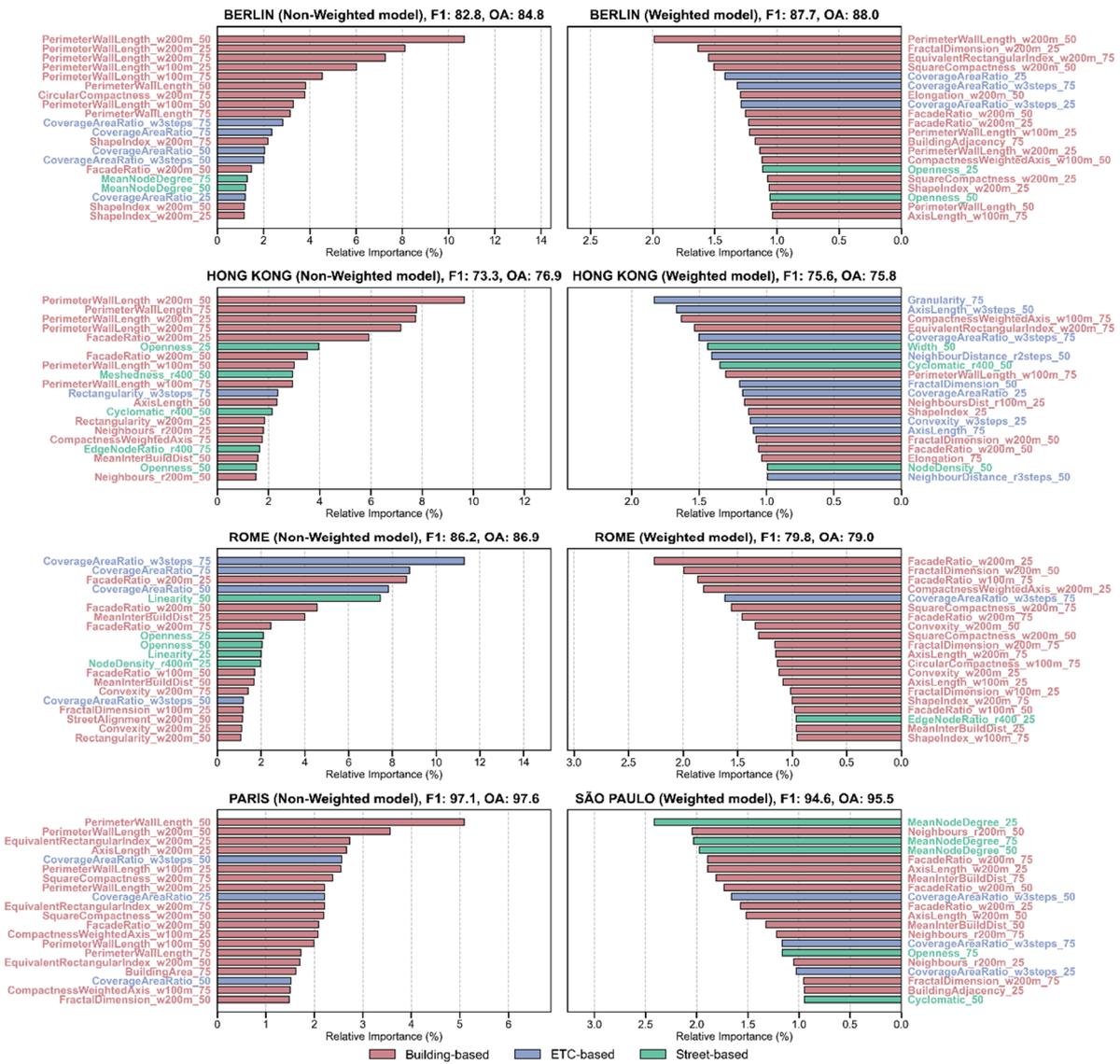

Figure F1. The subsets of 20 key morphometrics identified by the best-performing RF models, later used in fusion-based LCZ predictions.

# Supplementary material references